\documentclass{clv2-eng}

\usepackage{longtable}
\usepackage{graphicx}
\usepackage{comment}
\usepackage{url}
\usepackage{array}
\usepackage{tabularx}
\usepackage{booktabs}
\usepackage{caption}
\usepackage{adjustbox}

\makeatletter
\@ifundefined{numdef}{}{\let\numdef\relax}
\makeatother
\usepackage{etoolbox}

\issue{1}{1}{2006}

\title{Charting a Decade of Computational Linguistics in Italy: The CLiC-it Corpus}

\runningtitle{CLiC-it Corpus and Community Trends}

\runningauthor{Alzetta et al.}

\begin{document}

\author{Chiara Alzetta\thanks{Corresponding Author. E-mail:~\texttt{chiara.alzetta@ilc.cnr.it}}}
\affil{Istituto di Linguistica Computazionale ``Antonio Zampolli'', CNR, Pisa $-$ ItaliaNLP Lab}

\author{Serena Auriemma}
\affil{CoLingLab, Department of Philology, Literature and Linguistics, University of Pisa}

\author{Alessandro Bondielli}
\affil{Department of Computer Science, University of Pisa \newline CoLingLab, Department of Philology, Literature and Linguistics, University of Pisa}

\author{Luca Dini}
\affil{University of Pisa, Istituto di Linguistica Computazionale ``Antonio Zampolli'', CNR, Pisa $-$ ItaliaNLP Lab}

\author{Chiara Fazzone}
\affil{Istituto di Linguistica Computazionale ``Antonio Zampolli'', CNR, Pisa $-$ ItaliaNLP Lab}

\author{Alessio Miaschi}
\affil{Istituto di Linguistica Computazionale ``Antonio Zampolli'', CNR, Pisa $-$ ItaliaNLP Lab}

\author{Martina Miliani}
\affil{CoLingLab, Department of Philology, Literature and Linguistics, University of Pisa}

\author{Marta Sartor}
\affil{Istituto di Linguistica Computazionale ``Antonio Zampolli'', CNR, Pisa $-$ ItaliaNLP Lab}

\maketitle

\begin{abstract}
Over the past decade, Computational Linguistics (CL) and Natural Language Processing (NLP) have evolved rapidly, especially with the advent of Transformer-based Large Language Models (LLMs). This shift has transformed research goals and priorities, from Lexical and Semantic Resources to Language Modelling and Multimodality. In this study, we track the research trends of the Italian CL and NLP community through an analysis of the contributions to CLiC-it, arguably the leading Italian conference in the field. We compile the proceedings from the first 10 editions of the CLiC-it conference (from 2014 to 2024) into the CLiC-it Corpus, providing a comprehensive analysis of both its metadata, including author provenance, gender, affiliations, and more, as well as the content of the papers themselves, which address various topics. Our goal is to provide the Italian and international research communities with valuable insights into emerging trends and key developments over time, supporting informed decisions and future directions in the field.
\end{abstract}

\section{Introduction and Background}\label{intro}

The Italian Conference on Computational Linguistics (CLiC-it)\footnote{\url{https://www.ai-lc.it/en/conferences/clic-it/}} is the leading conference in Italy dedicated to Computational Linguistics (CL) and Natural Language Processing (NLP). It is organised by the Associazione Italiana di Linguistica Computazionale (AILC)\footnote{\url{https://www.ai-lc.it/en/}}, and it provides a platform for researchers, academics, and industry professionals to present advancements in NLP, linguistic resources, machine learning for language, and linguistic applications driven by Artificial Intelligence (AI).

The conference was first held in 2014, with its first edition in Pisa. Since then, it has been hosted by different Italian universities, including Trento, Napoli, Roma, Torino, Bari, Bologna, Milano, and Venezia. In 2024, for its tenth edition, CLIC-IT returned to Pisa, marking a decade of contributions to the Italian and international computational linguistics community.

\begin{table}[ht]
\centering
\resizebox{\columnwidth}{!}{%
\begin{tabular}{llllccc}
\hline
\textbf{Name} & \textbf{Date} & \textbf{Location} & \textbf{Chairs} & \multicolumn{1}{c}{\textbf{\begin{tabular}[c]{@{}c@{}}Papers \\ Received\end{tabular}}} & \multicolumn{1}{c}{\textbf{\begin{tabular}[c]{@{}c@{}}Papers\\ Accepted\end{tabular}}} & \multicolumn{1}{c}{\textbf{\begin{tabular}[c]{@{}c@{}}Acceptance \\ Rate\end{tabular}}} \\ \hline
CLiC-it 2014 & 9-11/12/2014 & Pisa & \begin{tabular}[c]{@{}l@{}}Roberto Basili, \\ Alessandro Lenci, \\ Bernardo Magnini\end{tabular} & 97 & 75 & 77.32 \\ \hline
CLiC-it 2015 & 3-4/12/2015 & Trento & \begin{tabular}[c]{@{}l@{}}Cristina Bosco, \\ Sara Tonelli, \\ Fabio Massimo Zanzotto\end{tabular} & 64 & 52 & 81.25 \\ \hline
CLiC-it 2016 & 5-6/12/2016 & Napoli & \begin{tabular}[c]{@{}l@{}}Anna Corazza, \\ Simonetta Montemagni, \\ Giovanni Semeraro\end{tabular} & 69 & 55 & 79.71 \\ \hline
CLiC-it 2017 & 11-13/12/2017 & Roma & \begin{tabular}[c]{@{}l@{}}Roberto Basili, \\ Malvina Nissim, \\ Giorgio Satta\end{tabular} & 72 & 58 & 80.56 \\ \hline
CLiC-it 2018 & 10-12/12/2018 & Torino & \begin{tabular}[c]{@{}l@{}}Elena Cabrio, \\ Alessandro Mazzei, \\ Fabio Tamburini\end{tabular} & 70 & 63 & 90.00 \\ \hline
CLiC-it 2019 & 13-15/11/2019 & Bari & \begin{tabular}[c]{@{}l@{}}Raffaella Bernardi, \\ Roberto Navigli, \\ Giovanni Semeraro\end{tabular} & 82 & 75 & 91.46 \\ \hline
CLiC-it 2020 & 1-3/03/2021 & \begin{tabular}[l]{@{}l@{}}Bologna\\ (online)\end{tabular} & \begin{tabular}[c]{@{}l@{}}Johanna Monti, \\ Felice Dell'Orletta, \\ Fabio Tamburini\end{tabular} & 80 & 69 & 86.25 \\ \hline
CLiC-it 2021 & 26-28/06/2022 & Milano & \begin{tabular}[c]{@{}l@{}}Elisabetta Fersini, \\ Marco Passarotti, \\ Viviana Patti\end{tabular} & 68 & 59 & 86.76 \\ \hline
CLiC-it 2023 & \begin{tabular}[c]{@{}l@{}}30/11 - \\ 02/12/2023\end{tabular} & Venezia & \begin{tabular}[c]{@{}l@{}}Federico Boschetti, \\ Gianluca E. Lebani, \\ Bernardo Magnini, \\ Nicole Novielli\end{tabular} & 86 & 75 & 87.21 \\ \hline
CLiC-it 2024 & 04-06/12/2024 & Pisa & \begin{tabular}[c]{@{}l@{}}Felice Dell’Orletta, \\ Alessandro Lenci, \\ Simonetta Montemagni, \\ Rachele Sprugnoli\end{tabular} & 136 & 114 & 83.82 \\ \hline
\end{tabular}%
}
\caption{Overview of CLiC-it conference editions.}\label{tab:clic-recap}
\end{table}

The conference has steadily grown over the years, both in terms of the number of papers and participants, as shown in Table \ref{tab:clic-recap}. It has become a staple of the Italian research communities for CL and NLP, fostering new contributions and ideas focused on the Italian language. With the rapidly evolving landscape of CL and NLP, driven by both increasing research volume and paradigm-shifting innovations such as Transformer-based Large Language Models (LLMs), new research directions have emerged globally and in Italy. The field has, for example, transitioned from a focus on lexical and semantic resources to topics such as large-scale language modelling, multimodal processing and applications such as affect detection and fact-checking. This transformation is visible not only on the international stage but also within the Italian research ecosystem, as reflected in CLiC-it.

In this study, we aimed to analyse and evaluate these ten years of CL and NLP in Italy through the lens of CLiC-it.
Several works have proposed similar evaluations of this research field, both at the national \cite{sprugnoli2019analisi,passaro2020lessons} and international level \cite{anderson-etal-2012-towards,mohammad2019state}.
Here, we attempt to move a step forward and provide the data that drove our analysis, in the form of the CLiC-it Corpus. %\footnote{\url{https://github.com/alemiaschi/clic-it_corpus}}. 
This corpus is a curated collection of ten years of conference proceedings. 

Beyond merely compiling papers, our work provides a detailed analysis of both metadata, such as author affiliations, geographic and gender distributions, and content, investigating how research topics have developed over time. The corpus offers a lens into how Italian NLP research has evolved.

The CLiC-it Corpus and the results of the analysis we carried out on the data, presented in this contribution, are a valuable resource for any NLP researcher interested in the Italian computational linguistics landscape, both long-standing and newcomers. 
It can be in fact particularly useful for those new to the field who wish to gain a broad overview of research trends, collaboration networks, and institutional contributions. At the same time, more experienced researchers may find it insightful to explore long-standing questions about authorship, international participation, and the evolution of topics within the Italian NLP community, supported by empirical evidence. Additionally, it is a valuable resource to keep track of the evolution of the Italian NLP community over time. 

Our work contributes to the NLP and CL research communities, both Italian and global, in the following ways:
\begin{itemize}
    \item We introduce and release the CLiC-it corpus, created by semi-automatically parsing all accepted papers across all editions of CLiC-it. The corpus is organised to provide core information on each research paper.
    \item We provide an in-depth analysis of the corpus, focusing both on metadata, such as distributions of participants, their affiliations and gender, their research age, the Italian networks formed over the years, etc., and on content, providing an analysis of the major topics that have interested the community over the years. 
\end{itemize}

This paper is organised as follows. Section \ref{data} introduces the CLiC-it corpus, detailing its structure and the criteria adopted for its creation. Section \ref{metadata} presents an in-depth analysis of the corpus metadata, focusing in particular on the authors of the papers presented at the conference editions (Section \ref{authors_analysis}), and on institutional collaborations and co-authorship networks (Section \ref{sec:affl}). In Section \ref{topic_detection}, we investigate the thematic content of CLiC-it papers, first outlining the topic modelling methodology used (Section \ref{topic_methodology}), and then discussing the research trends that emerge from the analysis (Section \ref{topic_result_and_disc}). Finally, Section \ref{concl} summarises the main findings.

\section{The CLiC-it Corpus}\label{data}

%693 scientific papers written either in English or Italian. 

%\textbf{CHIARA A. 15/05: Questa sezione per me è pronta. Manca solo sistemare i dettagli che ho lasciato in commento (miei dubbi o richieste di informazioni più precise su alcune affermazioni). }

The CLiC-it Corpus is a collection of metadata and textual content from the 693 papers presented at the CLiC-it conference over its 10 editions.  Its primary goal is to serve as an openly accessible, structured resource for investigating research trends and examining the development of the Italian NLP community over time. By offering a comprehensive overview of the community's scholarly production, the corpus enables longitudinal analyses of authorship, collaboration, and thematic focus. Furthermore, its design ensures long-term value, as it can be easily extended to include papers from future editions of the conference, thus supporting ongoing monitoring of the community's evolution.

To enable such diverse types of analyses, the corpus is designed to include both metadata and text content of the presented papers. This data was primarily acquired from the source files of the papers, kindly provided by the members of the AILC community who organised the various editions of the conference over the years (see Acknowledgements). The source files were shared in editable formats, such as LaTeX or proprietary Word documents, which allowed for the extraction of clean and structured text. However, for some editions (2014, 2018, 2019, and 2023), it was not possible to retrieve the original source files. In these cases, we extracted the paper text directly from the official conference proceedings available in PDF format. To obtain machine-readable text suitable for inclusion in the corpus, we tested different extraction strategies. For instance, we employed NotebookLM\footnote{\url{https://notebooklm.google.com/?}} to convert PDFs into editable text, followed by manual correction and refinement to ensure accuracy and consistency.

Once we obtained editable versions of all the conference papers, we proceeded with the extraction of both metadata and textual content. This step required different strategies depending on the format of the original files. For LaTeX documents, we implemented an automated workflow using the Python TexSoup library\footnote{\url{https://texsoup.alvinwan.com/}}, which allowed us to reliably extract key metadata fields, such as author names, paper titles, and affiliations, by targeting the corresponding fields in the standard conference templates. 
We also parsed the main LaTeX files to automatically extract the text of specific sections, namely the abstract, introduction, and conclusions, while removing all comments and LaTeX formatting tags to produce clean textual data. In contrast, for papers provided as Word files or as plain text obtained from PDF extraction, the metadata and relevant content were extracted manually. Since the CLiC-it conference allows the submission of papers both in English and Italian, we applied a further step to ensure linguistic uniformity across the corpus. All papers written in Italian (the total number across all editions is 47) were automatically translated into English using EasyNMT,\footnote{\url{https://github.com/UKPLab/EasyNMT}} a Python library for neural machine translation. 

Once this process was completed for all editions, we manually reviewed the extracted data to assess the consistency of data formats and apply necessary normalisation procedures. For author names, normalisation involved correcting orthographic variations, applying title case (i.e. capitalising only the first letters) when necessary, and standardising the order of given names and surnames. Institutional affiliations were also harmonised: university names were retained at the main institutional level, with references to internal departments or laboratories systematically omitted. In the case of national research centres, the full institute name was preserved.  Additionally, Italian institutions were consistently reported in Italian (e.g., Universit\'a di Firenze instead of University of Florence), while non-Italian institutions were recorded using their international names. Non-academic corporate names remained unaltered. 
As for the content sections, we focused on including only the abstract, introduction, and conclusion of each paper, as these are standard components across scientific publications and are consistently present in all submissions. This selection ensures comparability across papers and editions while retaining the most informative parts of each contribution.

The final dataset includes both the directly extracted content and metadata described above, as well as some additional information derived from them. Specifically, it provides details on the year of publication and the language in which each paper was written, alongside the names and affiliations of the authors. It also contains enriched data on authorship, such as the number of authors per paper, the number of women among them, and whether the first author is female. Institutional details are also reported, including the number of unique affiliations per paper, the presence of non-Italian institutions, and the number and names of authors affiliated with non-Italian and non-academic (corporate) entities.
Explicitly reporting this data enables a more nuanced understanding of the structure and dynamics of the CLiC-it research community. For instance, it makes the dataset adequate for quantitative analyses of gender representation, international collaboration, and the role of academic versus corporate research within the field. By making these variables readily available, the dataset serves as a valuable resource not only for investigating research trends and community evolution but also for supporting meta-scientific studies on scholarly practices in the Italian NLP landscape. 

The CLiC-it Corpus is freely available\footnote{Link to the github repository: \url{https://github.com/alemiaschi/clic-it_corpus}} and can be used for future studies aimed at further exploring these dimensions or integrating them with other national and international research initiatives.

\section{Contributors and Collaborations}\label{metadata}

%\textbf{CHIARA A. 22/05: Questa sezione per me è rivista. Tra i miei commenti più rilevanti segnalo la proposta di rifare alcuni grafici e alcune proposte (in commento o come bold nel testo) su come interpretare i risultati delle analisi. }

As a first step in our analysis of the CLiC-it corpus, we focus on the metadata associated with each paper, particularly the information relative to the paper authors and affiliations. This analysis aims to uncover trends and patterns in the composition of the AILC community over the ten editions of the CLiC-it conference, providing insights into both individual participation and institutional collaboration dynamics.

\subsection{Methodology} \label{metadata_methodology}

This section presents a two-part analysis, each tailored to address specific aspects of the metadata and guided by distinct methodological approaches.

Section \ref{authors_analysis} presents an in-depth analysis of the authors, with particular attention to the institutional affiliation of contributors. We distinguish between Italian and non-Italian institutions, as well as between academic and non-academic entities. We also track the number of new authors introduced in each conference edition. This part of the analysis relies on descriptive statistics and distributional trends to offer a diachronic perspective on the development of the CLiC-it author base and the extent of international participation.

Section \ref{sec:affl} shifts the focus to the structure and development of inter-institutional collaboration. To this end, we constructed a weighted directed graph of affiliations, $G=(V, E, W)$, where each node $v \in V$ represents an institution. A directed edge $e = (v_i, v_j)$ denotes that authors affiliated with $v_i$ have co-authored at least one paper with authors affiliated with $v_j$. The weight $w_{ij}$ assigned to each edge was computed following the method proposed by \cite{LIU20051462}, which balances both the frequency of collaboration and the relative contribution of each institution in multi-authored papers.
To this aim, we computed the weights of edges incorporating the following factors: \begin{enumerate} 
\item Institutions that collaborate more frequently should be assigned a higher collaboration weight; 
\item In papers with many authors from different institutions, the weight of each affiliation pair should be proportionally reduced. 
\end{enumerate}

This network-based approach allows us to analyse not only the volume but also the intensity and structure of collaborations within the CLiC-it community, offering a fine-grained view of how institutional partnerships have emerged and consolidated throughout the years, as described in Section \ref{sec:centrality}.

\subsection{Authorship Trends} \label{authors_analysis}

A research community is defined not only by its scientific outputs but also by the people who actively contribute to it. Understanding who contributes to a conference, how many, how often, and from where, provides valuable insights into the composition, growth, and inclusivity of that community. Tracking authorship over time captures patterns of continuity and renewal, the impact of outreach and institutional engagement.
%We begin by examining the number of authors contributing to each conference edition. This dimension provides insight into the size and evolution of the research community engaged with CLiC-it over time.
As a way of charting the development of the Italian NLP community, in this section we explore authorship trends across the ten editions of the CLiC-it conference. 

\begin{figure}
\begin{center}
\includegraphics[width=\textwidth]{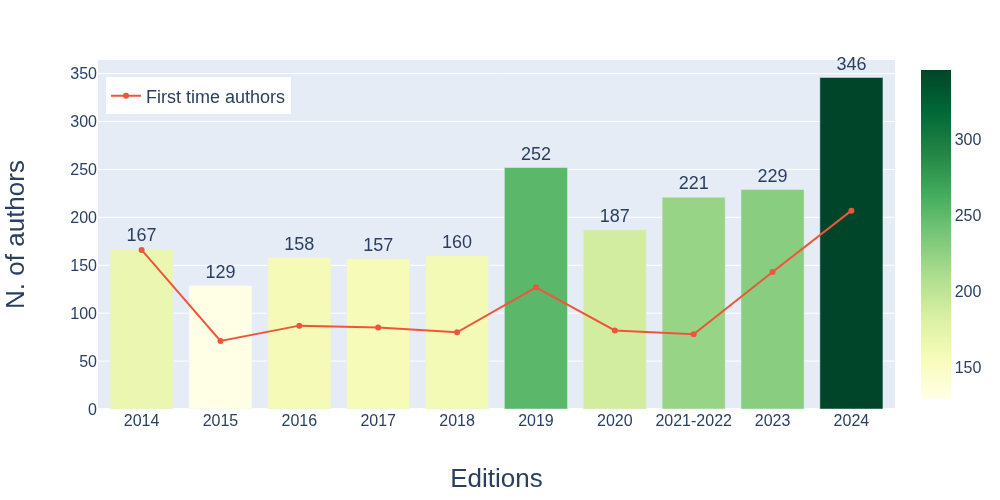}
\caption{Number of different authors in each edition, highlighting those submitting an article to the conference for the first time.}
\label{fig:authors}
\end{center}
\end{figure}

Over the course of the ten editions, a total of 2,006 unique authors contributed to the conference. %, while the overall number of authors, including repetitions, amounts to 2,348. This distinction reflects the fact that m
The inherently interdisciplinary nature of the CL and NLP fields, where research often emerges from collaborations across different areas of expertise, makes multi-authored publications the norm. The CLiC-it corpus reflects this collaborative dynamic, with an average of 3.39 authors per paper across all editions.
%Most papers have multiple authors, with an average of 3.39 authors per paper across all editions.
As shown in Figure \ref{fig:authors}, the number of different authors per edition shows significant variation, indicating shifts in participation and engagement over time. The inaugural 2014 edition attracted 167 authors, a strong starting point that confirmed the presence of a solid Italian NLP community and the need to create a venue to share and discuss their research. The following years saw a generally upward trend, with some fluctuations. A steady increase marked the 2019 edition, which registered 252 authors. This increase aligns with a broader wave of global interest in NLP, driven by major technological advances (e.g., BERT \cite{devlin2019bert} was released in 2018) and the growing relevance of language technologies across academia and industry. This momentum translated into higher participation in NLP conferences globally and likely influenced the national level as well, including CLiC-it \cite{mariani2022nlp4nlp+}. Despite the challenges posed by the COVID-19 pandemic, both the 2020 and 2021 editions continued to see substantial participation. The 2024 edition reached a record-breaking number of 346 authors, demonstrating the growing appeal and reach of the conference and its topics.

\begin{figure}
\begin{center}
\includegraphics[width=\textwidth]{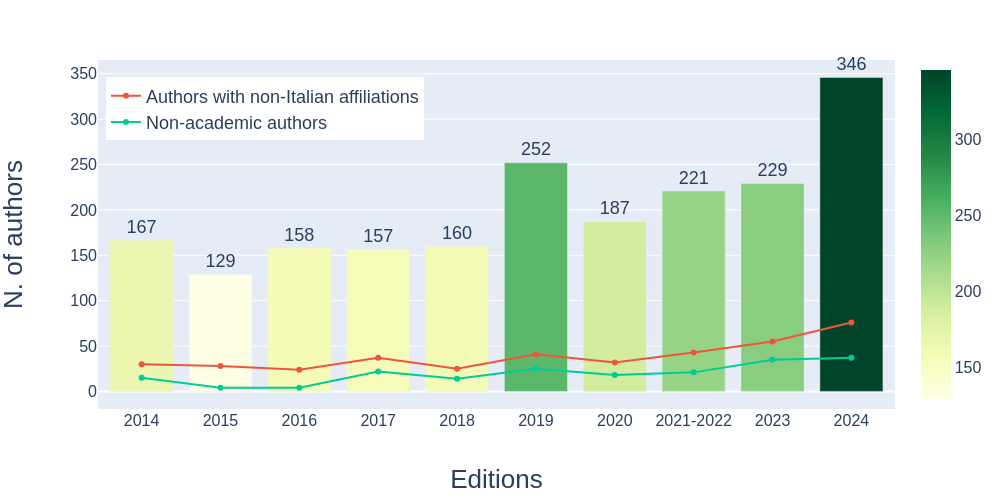}
\caption{Number of different authors in each edition, highlighting those with non-academic affiliations and with a non-Italian affiliation.}
\label{fig:authors_type}
\end{center}
\end{figure}

A particularly informative metric in understanding the community's dynamics is the number of first-time authors per edition. These are, for each edition, individuals who had never contributed to CLiC-it before, and are represented by the red line in Figure \ref{fig:authors}. This measure offers insight into the openness and renewal of the community. % In each edition, a significant portion of authors are participating for the first time, reflecting both the attractiveness of the conference to newcomers and the evolving composition of the field.
The sustained presence of newcomers alongside returning authors points to the success of CLiC-it in fostering an inclusive research environment and, above all, indicates a growing interest of the research community towards the topics of the conference. When we consider the ratio between the number of first-time authors and the number of unique authors for each edition, we observe that around half of the contributors are new in any given year (51.78\% on average, excluding the 2014 edition, where all authors are new contributors). This ratio remains high across editions, indicating that the community is growing and continually refreshed by new contributors, such as early-career researchers, students, and scholars from adjacent disciplines.

\begin{figure}
\begin{center}
\includegraphics[width=\textwidth]{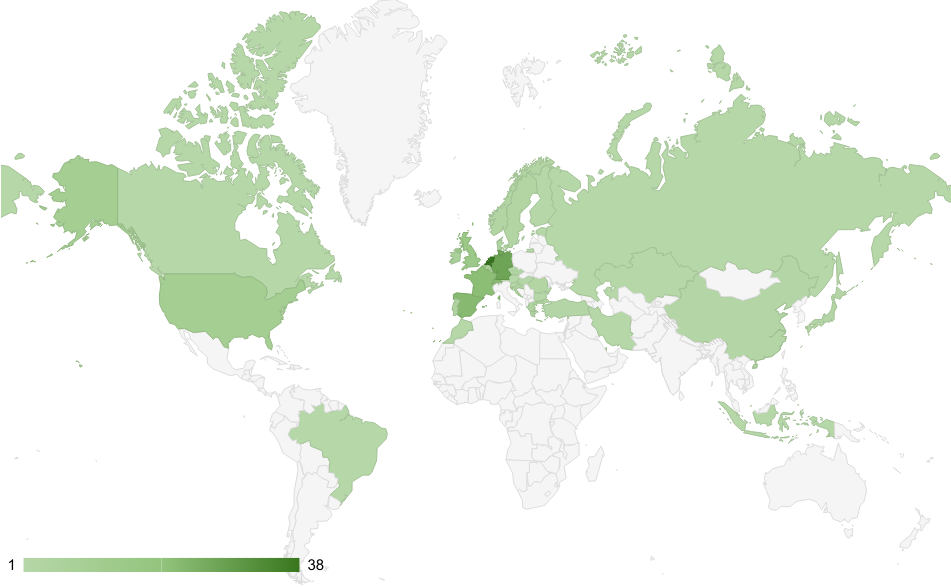}
\caption{Countries of non-Italian author affiliations submitting a paper at the CLiC-it conference. Darker colour shades reflect the number of papers submitted that include at least one author from that country.}
\label{fig:countries}
\end{center}
\end{figure}

Additionally, we focused on two subgroups: non-academic authors and authors with non-Italian affiliations. These categories capture the diversity and internationalisation of the community. Non-academic authors are those affiliated with companies, startups, or institutions outside the traditional academic and research domain (e.g. research and development in corporations) and are represented by the green line in Figure \ref{fig:authors_type}. Their presence, although smaller in absolute numbers, remained fairly stable across the years. This suggests a strengthening link between research and industry, and a rising interest in applied NLP from corporate and public-sector actors.
Similarly, the number of authors affiliated with non-Italian institutions (red line in Figure \ref{fig:authors_type}) provides a useful proxy for assessing the international reach and engagement of the CLiC-it conference. Although CLiC-it primarily targets the Italian computational linguistics community, its appeal has extended beyond national borders, attracting contributions from scholars based in institutions across the globe.
Figure \ref{fig:countries} illustrates the distribution of CLiC-it papers that include at least one author affiliated with a non-Italian institution. The map reveals a broad international participation, with authors from nearly every continent. Figure \ref{fig:countries_heatmap} in Appendix %~\ref{AppB}
B shows the distribution for each edition. The presence of international contributors, whether co-authors or lead authors, has increased gradually, contributing to the global visibility and relevance of the conference. In later editions, particularly from 2019 onwards, we observe a more consistent presence of international researchers. This is in line with the overall growing number of CLiC-it participants discussed above and proves the openness to cross-border collaborations of the Italian community.
When we consider the overall distribution in Figure \ref{fig:countries}, we notice that, as expected, European countries are the most prominently represented, reflecting both geographic proximity and strong academic ties with Italy. Notably, the Netherlands stands out as the most represented non-Italian country, with 38 papers involving Dutch-affiliated authors. This is followed by Germany (27 papers), and Spain and Switzerland, each contributing 22 papers. These results are likely influenced not only by geographic proximity but also by the high number of Italian researchers working in academic institutions within these countries.

More noteworthy, however, is the participation of authors affiliated with institutions in Asia and the Americas. While these regions are geographically distant from Italy, their presence in CLiC-it indicates the existence of international research collaborations, as we will discuss further in Section \ref{sec:affl}.

Together, these observations highlight the evolving role of CLiC-it as not only a national but also an increasingly international venue for scientific exchange in the field of computational linguistics. %point to a dynamic and expanding research ecosystem. 
The increasing number of contributors, the involvement of professionals outside academia, and the participation of international scholars all reflect the growth of the CLiC-it conference as a hub for NLP research and innovation.

\begin{figure}
\begin{center}
\includegraphics[width=\textwidth]{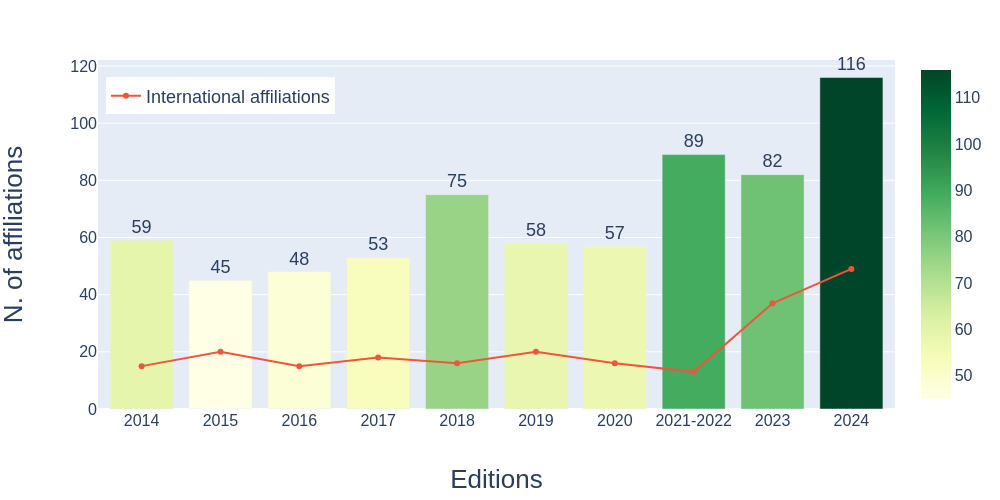}
\caption{Number of different affiliations in each edition, also showcasing the number of affiliations outside of Italy.}
\label{fig:affiliations}
\end{center}
\end{figure}

\subsection{Community Collaborations}\label{sec:affl}

\begin{figure}
    \centering
    \includegraphics[width=0.8\linewidth]{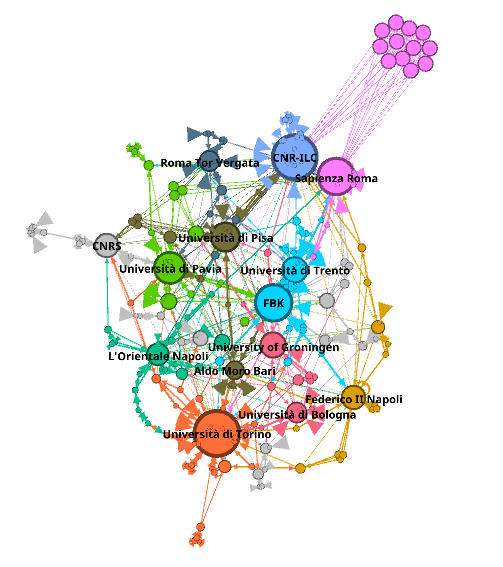}
    \caption{Giant Component in the network of collaborations between affiliations at CLiC-it.} %Giant Component of Affiliations Network.}
    \label{fig:affiliations_network}
\end{figure}

\begin{comment}

To examine the structure and evolution of institutional collaboration within the Clic-it conference over the past ten years, we constructed a weighted directed graph of affiliations, $G=(V, E, W)$. Each node $v \in V$ represents an institution, and a directed edge $e = (v_i, v_j)$ from institution $v_i$ to institution $v_j$ indicates that authors affiliated with $v_i$ have co-authored papers with authors affiliated with $v_j$. The weight $w_{ij}$ of each edge was computed following the methodology proposed in \cite{LIU20051462}, which accounts for both the frequency and strength of collaborative ties.

Specifically, the edge weights incorporate two key factors: \begin{enumerate} \item Institutions that collaborate more frequently should be assigned a higher collaboration weight; \item In papers with many authors from different institutions, the weight of each individual affiliation pair should be proportionally reduced. \end{enumerate}

\subsubsection{Network Properties} 
\end{comment}

%The network built using the data from the CLiC-it corpus and the methodology described in Sec. \ref{metadata_methodology} consists of 
The collaboration network constructed from the CLiC-it corpus, following the methodology outlined in Section \ref{metadata_methodology}, comprises 280 nodes and 1,350 edges. Each node represents a distinct institutional affiliation, while each edge corresponds to a collaboration, defined as the co-authorship of at least one paper by members of the connected institutions. The structure of this graph offers insights into the dynamics of institutional collaboration within the CLiC-it community.
%We performed a connected components analysis on the graph. Since $\forall e = (v_i, v_j) \in E$, there exists a corresponding edge $e' = (v_j, v_i) \in E$, every weakly connected component is also strongly connected. Thus, we refer to them simply as Connected Components (CCs).
To this aim, we performed a connected components analysis, a graph-based method used to identify and isolate subgraphs in which all nodes are connected to each other by at least one path.
Typically, the method allows distinguishing between weakly and strongly connected components depending on whether the direction of edges is respected.
The analysis involves traversing the graph using algorithms like depth-first search (DFS) or breadth-first search (BFS) to group nodes into their respective components, revealing the underlying structure and partitioning of the graph.
% Specifically, we computed the strongly connected components (SCCs) of the directed graph. However, d
Due to the symmetric nature of the edges in our graph ($\forall e = (v_i, v_j) \in E$, there exists a corresponding edge $e' = (v_j, v_i) \in E$), every weakly connected component is also strongly connected. This symmetry ensures that the direction of traversal does not impact the connectivity, making the graph undirected for the purposes of component analysis. Therefore, we use a standard depth-first search (DFS)-based or Union-Find algorithm to identify the components and refer to them simply as Connected Components (CCs). This approach allowed us to partition the graph into maximal subgraphs in which all nodes are mutually reachable, providing a clear structure for subsequent analysis.

%The graph contains 16 CCs, with a giant component comprising 233 nodes, represented in Figure \ref{fig:affiliations_network}. The remaining components are much smaller, with the largest containing only six nodes. 

The graph contains a total of 16 Connected Components (CCs), indicating that the network of affiliations is fragmented into several isolated subgroups. The largest of these is a giant component comprising 241 nodes, which is visually represented in Figure \ref{fig:affiliations_network}. This component reflects a dense core of institutions and affiliations that are interconnected through co-authorship or collaboration links. In contrast, the remaining 15 components are significantly smaller and more peripheral, with the largest among them including only six nodes, and many consisting of just a few or even isolated nodes. 

%Despite this large main component, the overall graph density remains low ($\approx 0.017$), indicating that most affiliations have never collaborated. The average node degree is 9.44, meaning that each affiliation has collaborated with approximately 9–10 others on average.

Despite the presence of this dominant component, the graph as a whole is sparsely connected, as reflected by the low overall graph density of approximately 0.017. Graph density, defined as the ratio of observed edges to the total number of possible edges in the network, is relatively low, indicating that collaborations tend to concentrate within specific, recurring clusters of institutions. The average node degree is 9.64, meaning that each affiliation has collaborated, on average, with 9 to 10 others. These figures suggest the presence of stable, long-term collaborations between institutions, likely driven by shared research interests and complementary expertise that have fostered consistent joint work over the years.

\subsubsection{Key Players in the Affiliation Network}\label{sec:centrality} 

%We conducted a centrality analysis to identify the most central affiliations in the collaboration network, using various measures to capture different aspects of centrality:

We conducted a centrality analysis to identify the institutional hubs and bridges in the collaboration network. Our goal is to gain insight into which institutions play prominent roles in the network, whether through direct connections or proximity to others. To this aim, we applied the following centrality measures to capture different dimensions of influence and connectivity. 

\begin{itemize}
    \item \textbf{Degree Centrality (DEG):} Degree centrality measures the number of edges directly connected to a node. In this setting, it represents the total number of unique collaborations an affiliation has engaged in. This metric reflects an affiliation's immediate activity level within the network and serves as the most straightforward indicator of prominence. Affiliations with high degree centrality are those that have collaborated with many others, suggesting broad engagement or a central role in multiple projects.
    %Measures the number of edges connected to a node. It captures how many direct collaborations an affiliation has and represents the most immediate and intuitive notion of centrality.

    \item \textbf{Closeness Centrality (CLS):} Closeness centrality captures how near a node is to all other nodes in the network by computing the reciprocal of the sum of the shortest path distances from that node to all others. This measure identifies affiliations that can quickly reach or influence others, either directly or through a few intermediaries. High closeness centrality indicates that an affiliation occupies a central position in the global structure of the network, making it more accessible and better positioned for the efficient spread of information or collaboration opportunities. %Reflects how close a node is to all others in the network. It is computed as the inverse of the sum of the shortest-path distances from the node to all others. Affiliations with high closeness centrality tend to have shorter paths to other institutions and can be seen as more "accessible."

    \item \textbf{Betweenness Centrality (BTW):} Betweenness centrality measures the extent to which a node lies on the shortest paths between other pairs of nodes. This metric reflects the potential of an affiliation to act as a bridge, facilitating interactions between otherwise disconnected parts of the network. Affiliations with high betweenness centrality can play a critical role in maintaining network cohesion and enabling indirect collaborations, even if their number of direct ties is relatively modest.
    %Quantifies how often a node lies on the shortest paths between pairs of other nodes. Affiliations with high betweenness centrality can be seen as key intermediaries or bridges within the network, potentially controlling the flow of information.
\end{itemize}

Each of these centrality measures complements the others by revealing different structural roles of affiliations. To obtain a global overview of the nodes' centrality, we created three rankings (i.e., one ranking for each metric) based on the decreasing centrality scores obtained by the graph nodes. Then, we computed the average ranking position of each node and created a global ranking of institutions based on their average ranking score. 
%To condense these different dimensions into a single indicator of centrality, we ranked all nodes separately for each of the three measures, assigning rank 1 to the institution with the highest score in a given centrality metric, rank 2 to the next, and so on. We then calculated the average rank for each node across the three centrality measures. This composite ranking provides 
This approach provides a more comprehensive view of centrality, capturing institutions that consistently hold prominent positions across multiple structural dimensions of the network. It also mitigates the potential bias of relying on any single centrality metric, which might emphasise only one aspect of connectivity.

% \paragraph{PageRank (PR)} A variant of eigenvector centrality that considers both the quantity and quality of connections. Affiliations that are connected to other well-connected nodes receive a higher score.

\begin{table}[t!] 
\centering 
\begin{tabular}{lrrrrr} 
\hline
    \textbf{Affiliation} & \textbf{DEG} & \textbf{CLS} & \textbf{BTW} & \textbf{Avg. Rank} & \textbf{Std Rank} \\ 
\hline
    Universit\'a di Torino & 0.34 & 0.42 & 0.20 & 1.33 & 0.58  \\
    Fondazione Bruno Kessler & 0.30 & 0.43 & 0.14 & 2.33 & 1.15 \\
    CNR-ILC & 0.34 & 0.41 & 0.16 & 2.67 & 1.15 \\
    Universit\'a di Pisa & 0.23 & 0.41 & 0.09 & 4.33 & 1.15 \\
    Sapienza Universit\'a di Roma & 0.30 & 0.38 & 0.10 & 5.00 & 2.65  \\
    Universit\'a di Pavia & 0.23 & 0.39 & 0.08 & 5.67 & 0.58 \\
    University of Groningen & 0.18 & 0.40 & 0.06 & 7.33 & 2.08 \\
    Universit\'a di Napoli Federico II & 0.16 & 0.38 & 0.06 & 9.33 & 1.15 \\
    CNRS & 0.16 & 0.37 & 0.07 & 9.33 & 2.08 \\
    Universit\'a di Napoli L'Orientale & 0.17 & 0.34 & 0.06 & 11.33 & 4.93 \\
\hline
\end{tabular} 
\caption{Centrality scores, average ranking positions with standard deviations for the 10 affiliations with the highest average centrality ranking.} %top affiliations.} 
\label{tab:affiliations_centrality} 
\end{table}

The results of the analysis are reported in Table \ref{tab:affiliations_centrality} and offer a view of institutional influence, from local engagement to global accessibility and intermediation. The University of Torino stands out as the most central institution across all measures, achieving the highest degree and closeness centrality, as well as a notably strong betweenness score, as indicated by its average ranking position of $1.33$ in the combined ranking. 
The standard deviation of the affiliation ranks is very low, with a maximum value of $4.93$. This indicates that the three centrality measures generally produce very similar rankings, suggesting a high level of consistency in how institutional importance is captured across different dimensions of network centrality.
%top average rank. 
This indicates that this institution is highly connected within the network. Fondazione Bruno Kessler and CNR-ILC follow closely, combining high connectivity (degree) with strong reachability (closeness), and moderate bridging roles (betweenness). Other Italian universities, such as Pisa, Roma Sapienza, and Pavia, exhibit balanced centrality scores, indicating active but slightly less dominant roles in the network. 

The presence of international institutions like the University of Groningen and CNRS further demonstrates the network's openness. Although their centrality scores are somewhat lower than those of the Italian core, their inclusion among the top 10 confirms their active involvement and structural relevance within the network.

%\textbf{Sono comunque nei top 10 per ranking globale, quindi proverei a mitigare un po' questa osservazione. In generale mi pare che tutte le ultime 4 affiliation della tabella siano carattezzate dal fatto che, pur essendo molto presenti nel grafo (quindi numerosi papers), collaborano poco con altre istituzioni (o forse sempre con le solite). Interpreto bene? Poi mi pare che i valori di CLS siano piuttosto stabili (c'è forse un drop su istituzioni più basse nel ranking?), mentre i valori di DEG e BTW tendono a scendere drasticamente nel ranking. Questo immagino sia coerente con il ranking per singola metrica... Cosa indica questo risultato rispetto alle tendenze delle istituzioni? Se volessimo approfondire ulteriormente si potrebbe vedere la presenza negli anni di queste istituzioni e capire se hanno cambiato il loro trend di comportamento da un anno all'altro.}
Overall, the table reveals a dense central cluster of affiliations that serve as both hubs and bridges, shaping the collaborative dynamics of the research community.

\section{Content Analysis} \label{topic_detection}

Having examined authorship patterns and institutional collaborations within the CLiC-it community, we now turn our attention to the scientific content of the conference. Understanding which topics have been most frequently addressed over the years provides a complementary perspective on the community’s evolution, highlighting shifts in research priorities and the emergence of new areas of interest.

This section outlines the methodology employed to analyse and categorise the conference papers published over the years, as described in Section \ref{topic_methodology}. The goal is to identify prevailing research topics and trends. The findings derived from this content analysis are presented and discussed in Section \ref{topic_result_and_disc}.

\begin{table}[t!]
\centering
\small
%\renewcommand{\arraystretch}{1.2}
%\resizebox{\textwidth}{!}{%
%\begin{tabular}{@{}l l c@{}}
\renewcommand{\arraystretch}{1.2}
\begin{tabular}{p{4cm}p{7cm}c}

\hline
\textbf{Topic Labels} & \textbf{Keywords} & \\
\hline
Language Models & language models, large language, natural language, multilingual, Italian language. \\
\hline
Chatbots and Dialogue Systems & dialogue systems, conversational agents, conversational agent, dialogues, chatbots. \\
\hline
Syntax and Dependency Treebanks & dependency treebank, dependency parsing, treebanks, parsers, annotation.\\
\hline
Latin Resources & latin texts, lexical resources, resources latin, linguistic resources, corpus. \\
\hline
Sentiment, Emotion, Irony, Hate & social media, sentiment analysis, affective, hate speech, irony detection. \\
\hline
Lexical and Semantic Resources and Analysis & text mining, computational lexicon, knowledge graph, morphological, frame semantics, annotation. \\
\hline
Learner Corpora and Language Acquisition & learner corpora, written language, linguistic, corpus, native language. \\
\hline
Textual Genres \& Literature Linguistics & textual genres, linguistic features, linguistic, literature. \\
\hline
Machine Translation & machine translation, automatic translation, translation systems, translation quality, human translation. \\
\hline
Text Simplification & lexical simplification, text simplification, parallel corpus, corpora, administrative texts. \\
\hline
Fact Checking and Fake News Detection & fact checking, fake news, propaganda, deceptive, misinformation, headlines. \\
\hline
Multimodal & multimodal corpus, linguistic features, utterances, speech, phonetics. \\
\hline
In-domain IR and IE & document classification, text retrieval, legal documents, information extraction, medical domain. \\
\hline
Gender and Inclusive Language Studies & grammatical gender, gender bias, speaker gender, fair language, inclusive language. \\
\hline
Distributional Semantics & distributional semantics, semantic relatedness, semantic space, word embeddings, word representation. \\
\hline
\end{tabular}%
%}
\caption{Topic labels, identified by ChatGPT, and keywords returned by c-TF-IDF algorithm from each cluster of the CLiC-it corpus.}
\label{tab:topics}
\end{table}

\subsection{Methodology} \label{topic_methodology}

For the topic modelling analysis of the papers presented at CLiC-it over the years, we employed the BERTopic framework proposed by \cite{grootendorst2022bertopic}. This framework consists of three primary steps. First, a Transformer-based model is utilised to extract text embeddings. Next, these embeddings are processed using the UMAP (Uniform Manifold Approximation and Projection for Dimension Reduction) algorithm \cite{mcinnes2018umap} to reduce dimensionality and optimise the clustering process, which is subsequently performed using the HDBSCAN (Hierarchical Density-Based Spatial Clustering of Applications with Noise) algorithm \cite{mcinnes2017hdbscan}. HDBSCAN extends the density-based DBSCAN algorithm by adopting a soft-clustering approach, allowing for the identification of outliers, i.e., texts that are considered noise or texts that could potentially belong to multiple clusters. Furthermore, HDBSCAN enables the parameterisation of cluster formation by specifying a minimum number of texts per cluster. Finally, keywords associated with each topic are identified through c-TF-IDF (class-based Term Frequency-Inverse Document Frequency), a modified version of the popular TF-IDF measure. Note that in the present scenario, the texts of documents belonging to the same topic are concatenated and treated as a single document for the computation of the inverse document frequency.

% Firstly, we extracted abstracts, introductions, and conclusions from all the papers in the proceedings of the CLiC-it editions from 2014 to 2024. Then, we automatically translated Italian papers into English by using EasyNMT, a Python library for neural machine translation. Each paper was annotated with the relative CLiC-it edition year, and BERTopic processed all the papers. 
We applied BERTopic \footnote{https://maartengr.github.io/BERTopic/index.html} to the abstracts, introductions, and conclusions of all papers in the CLiC-it corpus (see Section \ref{data}) to generate topic embeddings for our analysis. The dimensionality reduction via UMAP was carried out using the following parameters: n\_neighbors = 13, n\_components =2 (number of dimensions), min\_dist = 0.01, and metric = 'cosine'. As for the HDBSCAN algorithm, we used the Euclidean Distance and a minimum number of elements per cluster of 7. Documents identified as outliers (185) by HDBSCAN were reassigned to the most similar existing clusters to ensure full topic coverage. %We reassigned the papers in the outlier cluster to the other existing clusters. Finally, we obtained 24 clusters, each annotated with 10 keywords extracted with c-TF-IDF. We manually merged clusters with overlapping or similar keywords, and used ChatGPT to derive a single label for each cluster from these keywords (see Table \ref{tab:topics}). As a result, we divided the corpus in 15 clusters, as shown in Figure \ref{fig:topic_bar}.

This process initially resulted in 24 clusters, each described by 10 keywords generated with c-TF-IDF. We then manually merged clusters that showed substantial semantic overlap based on their keyword profiles. Finally, we used ChatGPT to generate a single descriptive label for each cluster, yielding a total of 15 refined clusters. % as illustrated in Figure \ref{fig:topic_plot} and detailed in Table \ref{tab:topics}.
Table \ref{tab:topics} reports the list of topics and keywords found using the methodology described above. 

\subsection{Research Trends in the CLiC-it Community}\label{topic_result_and_disc}

\begin{figure}
\begin{center}
\includegraphics[width=\textwidth]{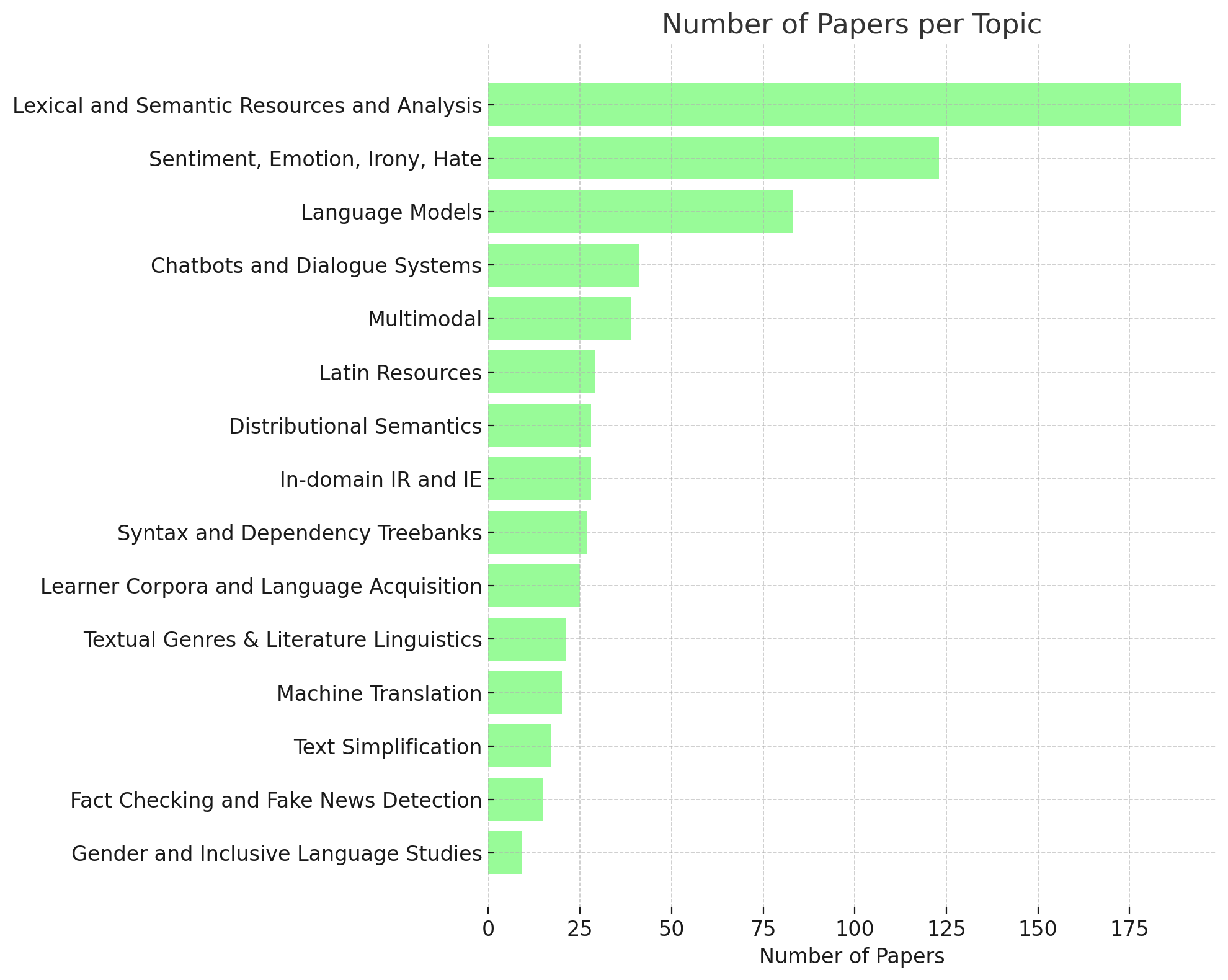}
\caption{Number of CLiC-it papers organised by topic.}
\label{fig:topic_plot}
\end{center}
\end{figure}

%Table \ref{tab:topics} presents the list of topics 
Figure \ref{fig:topic_plot} displays the distribution of papers across the 15 thematic clusters identified in the CLiC-it corpus through topic modelling. Overall, the clusters reflect both methodological approaches and domain-specific applications of NLP and computational linguistics.

The most represented topic is \textit{Lexical and Semantic Resources and Analysis} (189 papers). The keywords extracted from the papers in this cluster indicate an interest towards resource creation (see keywords like `annotation', `morphological', `frame semantics') and analysis (`text mining', `computational lexicon', possibly referring to computational approaches to lexicography), as well as information representation approaches, suggested by the keyword `knowledge graph'. In addition to this, several other clusters represent foundational areas of NLP. The \textit{Syntax and Dependency Treebanks} cluster focuses on dependency parsing and syntactically annotated resources such as treebanks. The \textit{In-domain IR and IE} cluster addresses text retrieval and classification, as well as domain-specific information extraction techniques, particularly in the legal and medical domains. Finally, the \textit{Multimodal} cluster highlights the integration of textual data with speech and other modalities, as NLP systems evolve to handle richer and more complex input types. A high number of papers also fall under the \textit{Sentiment, Emotion, Irony, Hate} cluster (123 papers). The associated keywords indicate a strong focus of these papers on affective computing and ethical concerns, especially in the context of digital discourse. Such a result is likely driven by the widespread availability of social media data in past years and the popularity of their analysis within the NLP community. 

More specialised areas, including \textit{Latin Resources}, \textit{Learner Corpora} and \textit{Textual Genre and Literature Linguistics}, also emerge with a notable presence, indicating the diversity of the domains addressed at CLiC-it. These clusters partially overlap in theme with the larger cluster on lexical and semantic resources: while all discuss corpora of textual linguistic resources, the learner corpora and Latin resources papers seem to be more concerned with corpus construction and description rather than corpus analysis. Conversely, papers on literary linguistics discuss approaches for analysing the text style, as suggested by the keyword `linguistic features'. In between, \textit{Text Simplification} emerges as a topic addressed by a small portion of the CLiC-it community, although it addresses both issues related to corpus construction (see keywords like `parallel corpus' and `administrative texts') and to simplification approaches (`lexical simplification').

Finally, clusters like \textit{Fact Checking and Fake News Detection} and \textit{Gender and Inclusive Language Studies}, though smaller in size, point to the community's engagement with socially relevant research topics. In fact, while the former group is interested in detecting deceptive information from news corpora, the latter addresses issues related to gender biases through different data sources.

Topics such as \textit{Language Models} 
and \textit{Chatbots and Dialogue Systems} likely emerged more recently, indicating a shift toward neural and conversational technologies, consistent with broader trends in NLP. To explore how research interests have evolved over time, we performed a diachronic analysis of the topic distribution across CLiC-it editions. The results, displayed in Figure \ref{fig:topic_plot}, reveal several meaningful patterns and shifts in focus within the community. In what follows, we describe the most salient trends, highlighting the rise and decline of specific themes and relating them to broader developments in NLP research.

Interestingly, several topic clusters emerging from the CLiC-it corpus align closely with broader trends identified in large-scale analyses of the ACL Anthology \cite{mohammad2019state} spanning from the 1960s to 2020, although the two datasets overlap for a relatively short period (2014–2020). For example, the analysis of the ACL corpus reveals %both studies highlight 
the historical centrality of topics such as \textit{Machine Translation}, \textit{Sentiment Analysis}, and \textit{Parsing}, %. These areas were among the most frequent in ACL paper titles from the 1980s through the early 2000s, and their presence is 
also reflected in the CLiC-it proceedings.
Moreover, both datasets capture a growing interest in neural approaches for language modelling tasks. In the ACL corpus, this is evidenced by the rising frequency of terms such as \textit{neural}, \textit{word embeddings}, and \textit{deep learning}. This trend is mirrored in the CLiC-it corpus through the growing prominence of the topic \textit{Language Models} (see figure \ref{fig:topic_plot}), where keywords such as `large language' and `language models' are likely indicative of the increasing use of large-scale and generative neural models. Finally, both corpora reveal a diversification of research topics over time. In the ACL Anthology, the dominance of a few key terms has diminished in recent decades in favour of a broader array of specialised areas. This evolution is echoed in the presence of niche clusters within the more recent CLiC-it corpus, such as \textit{Gender and Inclusive Language Studies}, and \textit{Fact Checking and Fake News Detection}.

\subsubsection{Diachronic Analysis of Topics}

\begin{figure}[ht]
\begin{center}
\includegraphics[width=\textwidth]{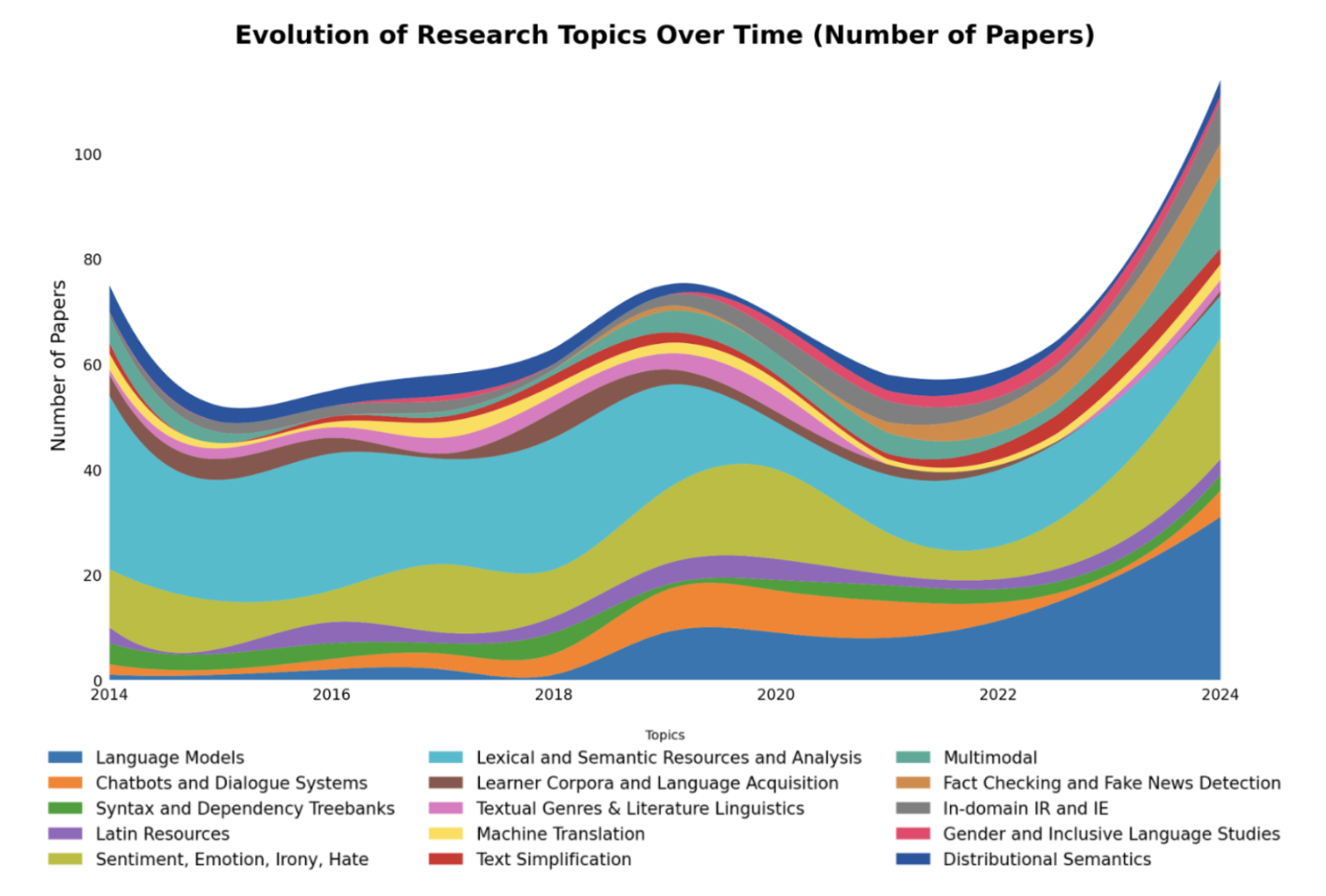}
\caption{Coverage of topics in CLiC-it papers across editions.}
\label{fig:topic_plot}
\end{center}
\end{figure}

The evolution of research topics across CLiC-it conference editions reveals significant trends and shifts in focus within the Italian computational linguistics community. Figure \ref{fig:topic_plot} illustrates how these thematic clusters have varied over time, offering insight into both long-standing interests and emerging areas.

The earliest editions of CLiC-it (2014–2018) display a relatively stable distribution of topics, with a strong emphasis on resource-oriented research. In particular, the cluster on lexical and semantic resources dominates this period, reflecting the community's foundational efforts in building and analysing linguistic resources. Similarly, other resource-related clusters, including those addressing application scenarios of corpus analysis, feature prominently, indicating the centrality of corpus-driven approaches in the initial stages of the conference.

Among the most popular application-oriented topics in the early years was sentiment analysis, largely supported by the availability of social media data and the popularity of sentiment analysis methods. While this topic has been consistently relevant throughout the years, a noticeable dip occurred in 2018 and again in 2022. Conversely, a peak in 2020 may be attributed to increased online activity during the COVID-19 pandemic, which brought renewed attention to the affective and ethical dimensions of digital discourse.

Around this same period, papers focusing on literature analysis gradually diminished, giving way to growing interest in fake news detection and gender bias studies. These began with studies on traditional models such as Word2Vec around 2017, and evolved from 2019 onward to include Transformer-based approaches. This change in focus might reflect a growing interest within the community in the socio-cultural application of their research.

Also emerging in this transitional period (2019-2022) is the cluster on chatbots and dialogue systems.  %garnered attention from the NLP community. 
The models proposed by the papers falling into this topic reflect the shift from rule-based systems to dialogue systems based on more recent neural models. In fact, while most of the early proposed solutions were based on templates and graph databases, more recently, several resources have been produced to train conversational agents, enabling the development of more advanced data-driven models. As observed from Figure \ref{fig:topic_plot}, dialogue systems were quickly supplanted in 2022 by generative AI models based on LLMs.
%\textbf{COMMENTO CHIARA: questo topic su chatbot rimane misterioso perchè le keywords non permettono di capire se c'è un interesse verso le tecniche per la costruzione di chatbot o se questi chatbot sono proposti in qualche contesto particolare che può legarsi ai bias e al fact checking. Se vado a memoria ai lavori proposti a clic, penso che sia il secondo caso, ma andrebbe verificato con i paper finiti in questo cluster. }
%\textbf{COMMENTO Martina: chatbot e fact checking non hanno molto a che fare. ho risolto  guardando un po' ai paper e formulando il testo in corsivo.}

Some areas of research have maintained a steady presence throughout the years. Topics such as multimodality, Machine Translation, syntax, and Latin Resources have shown consistent contributions, underlining the community's commitment to traditional linguistic tasks and classical languages. In contrast, topics like Distributional Semantics, Information Retrieval and Extraction, and corpus construction have experienced a marked decline in recent years. This reduction appears to coincide with the rise of research on Language Models, which has seen substantial growth since 2018. The increasing focus on this topic reflects a broader shift in NLP towards pre-trained neural architectures and their application in multilingual and domain-specific contexts, with a particular emphasis on the Italian language.

The trajectory of the Language Models topic offers a clear example of the field's methodological evolution. Before 2018, papers largely employed traditional machine learning and deep learning architectures, including convolutional neural networks (CNNs), recurrent neural networks (RNNs), long short-term memory networks (LSTMs), and support vector machines (SVMs). From 2019 onwards, however, the Transformer architecture became predominant, marking a turning point in NLP research. Studies increasingly explored the fine-tuning of large pre-trained models for downstream tasks such as question answering and text classification. Notably, this phase saw the emergence of Italian-specific models such as GePpeTto \cite{de2020geppetto} and BERTino \cite{muffo2020bertino}, along with growing interest in transfer learning techniques and the evaluation of language models through metrics such as perplexity.

Looking forward, some currently declining topics may regain momentum in light of these recent technological developments. For instance, interest in Language Acquisition could be revitalised by the rise of Baby LLMs that target developmental linguistics and incremental learning. Similarly, the topic of text simplification may experience renewed relevance given the ability of generative models to transform and adapt texts for clarity and accessibility. 
Multimodality represents an example of long-standing areas that are experiencing growing attention. Although present since the first editions, multimodal research has expanded in recent years, reflecting increasing interest in integrating linguistic data with other communicative modalities, especially in the domain of spoken language analysis.

Overall, the diachronic analysis reveals a dynamic and responsive research community. From foundational linguistic resources to state-of-the-art generative models, the CLiC-it community has demonstrated a sustained ability to integrate continuity with innovation. Some areas have matured and stabilised, while others have emerged in response to new technological affordances and societal needs. This trajectory underscores the dual focus of the community: advancing methodological frontiers while addressing the ethical, inclusive, and application-driven dimensions of NLP.

\section{Conclusions}\label{concl}

In this study, we presented the CLiC-it Corpus, a structured and curated dataset encompassing ten years of research output from the Italian Conference on Computational Linguistics (CLiC-it). To assemble the corpus, a combination of automatic and manual extraction methods was used. In the end, we compiled metadata and textual content from a total of 693 scientific papers, offering a detailed snapshot of the Italian CL and NLP community from 2014 to 2024. Our analysis explored key dimensions of the community’s evolution, including authorship patterns, gender and institutional diversity, collaboration networks, and shifting research priorities.

The results of our investigation highlight a growing research landscape. We observed an increasing number of submissions and participants, a gradual internationalisation of contributions, and a diversification of research topics in line with global trends, such as the emergence of multimodality, Large Language Models, and socially impactful applications like misinformation detection and sentiment analysis. The analysis of co-authorship and institutional networks further illustrates the collaborative nature of the Italian NLP landscape and its openness to cross-institutional and interdisciplinary research.

By releasing the CLiC-it Corpus, we provide a valuable resource for researchers and practitioners, both Italian and foreign, interested in understanding the dynamics of NLP research in Italy. 
The corpus supports additional empirical studies on authorship, topic evolution, and collaboration, and can serve as a reference for future evaluations of the field’s trajectory. Our end goal was to facilitate more informed, inclusive, and strategically guided research efforts, both within Italy and in dialogue with the broader international NLP community.

The corpus is designed to be extensible and sustainable, and we expect to maintain it by including new research papers for upcoming editions of CLiC-it. Future updates will also enable longitudinal monitoring of the community’s development. We encourage further use and enrichment of the resource, and we hope it will inspire similar initiatives in other linguistic and national contexts.

\section*{Acknowledgments}

We thank the organisers of all past editions of the Italian Conference on Computational Linguistics for providing us with source papers and materials for assembling the CLiC-it corpus. 

\bibliographystyle{fullname}
\bibliography{bibliography}

\begin{thebibliography}{}

\bibitem[\protect\citename{Anderson, Jurafsky, and McFarland}2012]{anderson-etal-2012-towards}
Anderson, Ashton, Dan Jurafsky, and Daniel~A. McFarland.
\newblock 2012.
\newblock Towards a computational history of the {ACL}: 1980-2008.
\newblock In Rafael~E. Banchs, editor, {\em Proceedings of the {ACL}-2012 Special Workshop on Rediscovering 50 Years of Discoveries}, pages 13--21, Jeju Island, Korea, July. Association for Computational Linguistics.

\bibitem[\protect\citename{De~Mattei \bgroup et al.\egroup }2020]{de2020geppetto}
De~Mattei, Lorenzo, Michele Cafagna, Felice Dell’Orletta, Malvina Nissim, Marco Guerini, Aptus AI, and Fondazione~Bruno Kessler.
\newblock 2020.
\newblock Geppetto carves italian into a language model.
\newblock {\em Computational Linguistics CLiC-it 2020}, page 136.

\bibitem[\protect\citename{Devlin \bgroup et al.\egroup }2019]{devlin2019bert}
Devlin, Jacob, Ming-Wei Chang, Kenton Lee, and Kristina Toutanova.
\newblock 2019.
\newblock Bert: Pre-training of deep bidirectional transformers for language understanding.
\newblock In {\em Proceedings of the 2019 conference of the North American chapter of the association for computational linguistics: human language technologies, volume 1 (long and short papers)}, pages 4171--4186.

\bibitem[\protect\citename{Grootendorst}2022]{grootendorst2022bertopic}
Grootendorst, Maarten.
\newblock 2022.
\newblock Bertopic: Neural topic modeling with a class-based tf-idf procedure.
\newblock {\em arXiv preprint arXiv:2203.05794}.

\bibitem[\protect\citename{Liu \bgroup et al.\egroup }2005]{LIU20051462}
Liu, Xiaoming, Johan Bollen, Michael~L. Nelson, and Herbert {Van de Sompel}.
\newblock 2005.
\newblock Co-authorship networks in the digital library research community.
\newblock {\em Information Processing \& Management}, 41(6):1462--1480.
\newblock Special Issue on Infometrics.

\bibitem[\protect\citename{Mariani \bgroup et al.\egroup }2022]{mariani2022nlp4nlp+}
Mariani, Joseph, Gil Francopoulo, Patrick Paroubek, and Fr{\'e}d{\'e}ric Vernier.
\newblock 2022.
\newblock Nlp4nlp+ 5: The deep (r) evolution in speech and language processing.
\newblock {\em Frontiers in Research Metrics and Analytics}, 7:863126.

\bibitem[\protect\citename{McInnes \bgroup et al.\egroup }2017]{mcinnes2017hdbscan}
McInnes, Leland, John Healy, Steve Astels, et~al.
\newblock 2017.
\newblock hdbscan: Hierarchical density based clustering.
\newblock {\em J. Open Source Softw.}, 2(11):205.

\bibitem[\protect\citename{McInnes, Healy, and Melville}2018]{mcinnes2018umap}
McInnes, Leland, John Healy, and James Melville.
\newblock 2018.
\newblock Umap: Uniform manifold approximation and projection for dimension reduction.
\newblock {\em arXiv preprint arXiv:1802.03426}.

\bibitem[\protect\citename{Mohammad}2019]{mohammad2019state}
Mohammad, Saif~M.
\newblock 2019.
\newblock The state of nlp literature: A diachronic analysis of the acl anthology.
\newblock {\em arXiv preprint arXiv:1911.03562}.

\bibitem[\protect\citename{Muffo and Bertino}2020]{muffo2020bertino}
Muffo, Matteo and Enrico Bertino.
\newblock 2020.
\newblock Bertino: an italian distilbert model.
\newblock {\em Computational Linguistics CLiC-it 2020}, page 317.

\bibitem[\protect\citename{Passaro \bgroup et al.\egroup }2020]{passaro2020lessons}
Passaro, Lucia~C, Maria Di~Maro, Valerio Basile, and Danilo Croce.
\newblock 2020.
\newblock Lessons learned from evalita 2020 and thirteen years of evaluation of italian language technology.
\newblock {\em IJCoL. Italian Journal of Computational Linguistics}, 6(6-2):79--102.

\bibitem[\protect\citename{Sprugnoli \bgroup et al.\egroup }2019]{sprugnoli2019analisi}
Sprugnoli, Rachele, Gabriella Pardelli, Federico Boschetti, et~al.
\newblock 2019.
\newblock Un’analisi multidimensionale della ricerca italiana nel campo delle digital humanities e della linguistica computazionale.
\newblock {\em Umanistica Digitale}, (5):59--89.

\end{thebibliography}

\appendix
\appendixsection{Fields in the CLiC-it Corpus}

\begin{itemize}
    \item \textit{id}: A unique identifier assigned to each paper.
    \item \textit{year}: The year in which the paper was presented at the CLiC-it conference, useful for tracking trends over time.
    \item \textit{authors}: The names of all contributing authors. %, providing insights into collaboration networks.
    \item \textit{title}: The title of the paper.
    \item \textit{language}: The language in which the paper was written (ENG for English, ITA for Italian).
    \item \textit{num\_authors}: The total number of authors, indicating collaboration size and research team structure.
    \item \textit{num\_women\_authors}: The number of female authors, enabling gender distribution analysis within research contributions.
    \item \textit{woman\_as\_first\_author}: A binary indicator (1 or 0) reflecting whether the first author is a woman. %, supporting studies on gender representation in leading research roles.
    \item \textit{affiliations}: The institutional affiliations of the authors. %, shedding light on the diversity and reach of contributing institutions.
    \item \textit{num\_affiliations}: The number of distinct affiliations. %, measuring institutional collaboration in the research.
    \item \textit{at\_least\_one\_international\_affiliation}: A binary value (1 or 0) denoting whether at least one author is affiliated with a non-Italian institution.
    \item \textit{international\_affiliation\_only}: A binary value (1 or 0) indicating whether all authors belong exclusively to international (non-Italian) institutions. 
    \item \textit{international\_authors}: The number of authors with non-Italian affiliations.
    \item \textit{names\_international\_authors}: The names of authors with non-Italian affiliations.
    \item \textit{num\_company\_authors}: The number of authors affiliated with corporate (non-academic) institutions, revealing industry engagement in research.
    \item \textit{names\_company\_authors}: The names of authors working in corporate settings.
    \item \textit{countries\_affiliations}: The countries represented by the affiliations, providing a geographical scope of institutional participation.
    \item \textit{cities\_affiliations}: The cities of the affiliations, adding a finer-grained geographical perspective to the institutional landscape.
    \item \textit{abstract}: The paper abstract.
    \item \textit{introduction}: The paper Introduction section.
    \item \textit{conclusion}: The paper Conclusions section.
    \item \textit{num\_topics}: Number of topics extracted with BERTopic.
    \item \textit{topics}: The topics extracted with BERTopic.
\end{itemize}

\appendixsection{Participation to CLiC-it by Country} \label{AppB}

\begin{tabular}{lrrrrrrrrrr}
\hline
 & 2014 & 2015 & 2016 & 2017 & 2018 & 2019 & 2020 & 2021 & 2023 & 2024 \\
\hline
Italy & 67 & 46 & 54 & 51 & 58 & 68 & 64 & 52 & 70 & 100 \\
Austria & 0 & 0 & 0 & 0 & 0 & 1 & 0 & 0 & 0 & 0 \\
Belgium & 0 & 0 & 0 & 0 & 0 & 1 & 1 & 3 & 1 & 1 \\
Brazil & 0 & 0 & 0 & 0 & 0 & 0 & 0 & 1 & 0 & 0 \\
Bulgaria & 0 & 0 & 0 & 0 & 0 & 0 & 0 & 0 & 1 & 0 \\
Canada & 0 & 0 & 1 & 0 & 0 & 0 & 0 & 0 & 0 & 0 \\
China & 1 & 0 & 0 & 0 & 0 & 0 & 0 & 1 & 0 & 0 \\
Croatia & 0 & 0 & 0 & 2 & 1 & 1 & 0 & 0 & 0 & 0 \\
Czech Republic & 0 & 0 & 0 & 0 & 0 & 0 & 1 & 1 & 1 & 1 \\
Denmark & 0 & 0 & 1 & 0 & 0 & 0 & 0 & 0 & 1 & 2 \\
Estonia & 0 & 0 & 0 & 0 & 0 & 0 & 0 & 0 & 1 & 0 \\
Finland & 0 & 0 & 0 & 0 & 0 & 0 & 0 & 0 & 0 & 1 \\
France & 3 & 2 & 1 & 1 & 6 & 1 & 0 & 1 & 1 & 5 \\
Germany & 1 & 3 & 1 & 3 & 4 & 4 & 0 & 1 & 3 & 7 \\
Greece & 1 & 0 & 1 & 0 & 0 & 0 & 0 & 1 & 1 & 0 \\
Hungary & 0 & 0 & 0 & 0 & 1 & 0 & 0 & 0 & 1 & 0 \\
Indonesia & 0 & 0 & 0 & 0 & 0 & 0 & 0 & 0 & 1 & 0 \\
Iran & 0 & 1 & 0 & 0 & 0 & 0 & 0 & 0 & 0 & 0 \\
Ireland & 0 & 2 & 1 & 0 & 0 & 2 & 2 & 0 & 0 & 0 \\
Japan & 0 & 0 & 0 & 0 & 0 & 0 & 0 & 0 & 1 & 1 \\
Kazakhstan & 1 & 0 & 0 & 0 & 0 & 0 & 0 & 0 & 0 & 0 \\
Luxembourg & 0 & 1 & 0 & 0 & 0 & 0 & 0 & 0 & 0 & 0 \\
Malta & 0 & 0 & 0 & 1 & 0 & 0 & 1 & 0 & 0 & 0 \\
Morocco & 0 & 0 & 0 & 0 & 0 & 0 & 0 & 0 & 0 & 1 \\
Netherlands & 3 & 4 & 3 & 3 & 4 & 5 & 7 & 2 & 3 & 4 \\
Norway & 0 & 1 & 0 & 0 & 0 & 0 & 1 & 1 & 0 & 0 \\
Portugal & 0 & 0 & 0 & 0 & 0 & 0 & 0 & 0 & 1 & 3 \\
Qatar & 4 & 1 & 2 & 3 & 0 & 0 & 0 & 0 & 0 & 0 \\
Romania & 0 & 0 & 0 & 0 & 0 & 0 & 1 & 0 & 0 & 2 \\
Russia & 1 & 0 & 0 & 0 & 0 & 0 & 0 & 0 & 0 & 0 \\
Singapore & 0 & 1 & 0 & 0 & 0 & 0 & 1 & 0 & 0 & 0 \\
Slovenia & 0 & 0 & 0 & 0 & 0 & 0 & 0 & 0 & 1 & 0 \\
Spain & 1 & 1 & 1 & 3 & 1 & 2 & 1 & 1 & 5 & 6 \\
Sweden & 0 & 0 & 1 & 1 & 0 & 0 & 0 & 0 & 0 & 1 \\
Switzerland & 0 & 1 & 1 & 3 & 0 & 3 & 1 & 2 & 7 & 4 \\
Turkey & 0 & 0 & 0 & 0 & 0 & 1 & 1 & 0 & 0 & 0 \\
United Kingdom & 1 & 2 & 2 & 1 & 0 & 0 & 0 & 0 & 3 & 6 \\
USA & 3 & 1 & 1 & 0 & 0 & 1 & 1 & 0 & 2 & 1 \\
\hline
\end{tabular}

\begin{figure}
\begin{center}
\includegraphics[width=\textwidth]{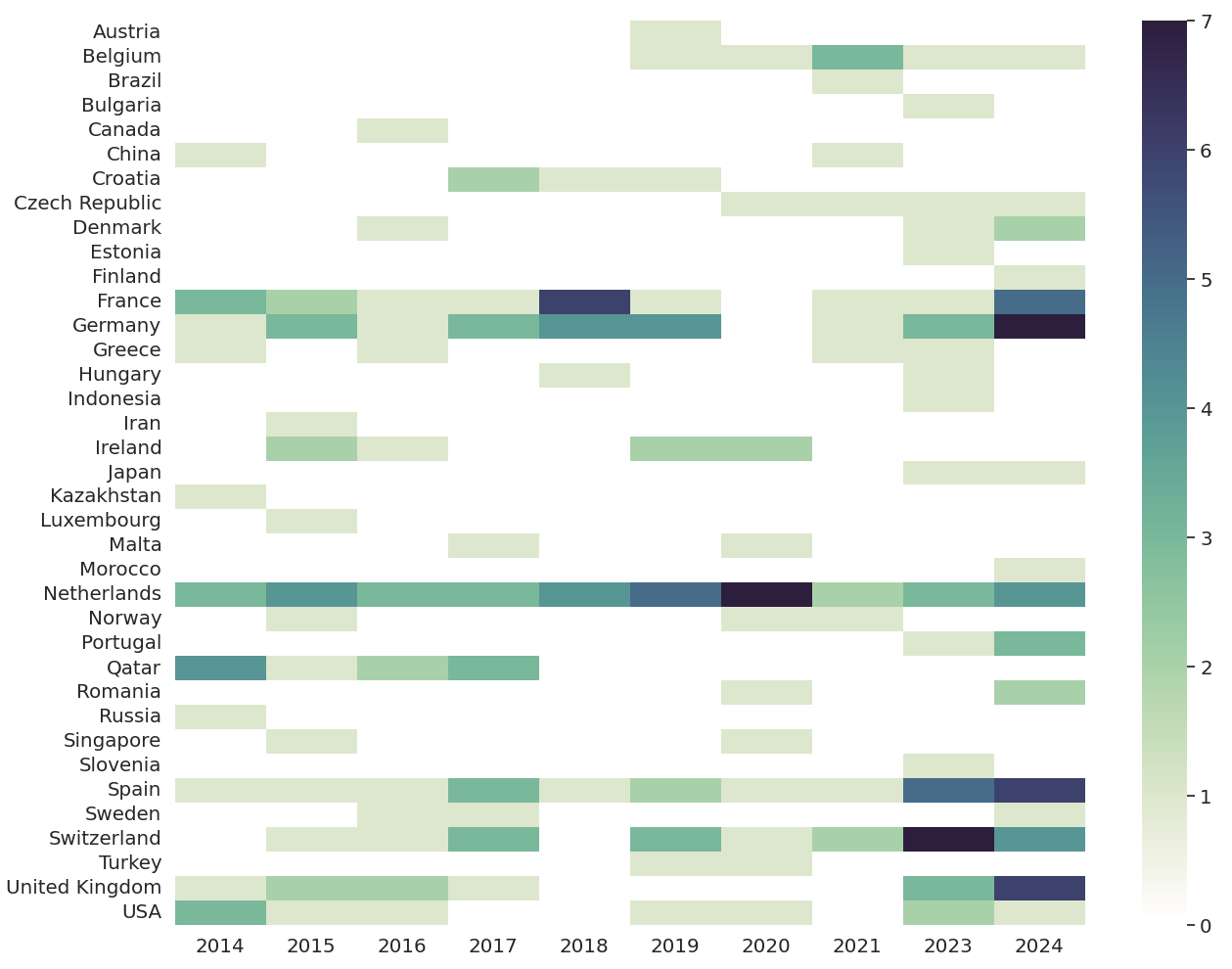}
\caption{Number of affiliations from each country (besides Italy) per edition.}
\label{fig:countries_heatmap}
\end{center}
\end{figure}

\end{document}